\documentstyle[twoside,epsf,12pt]{kaeog}

\newfont{\mm}   {msbm10}              % mathematical upper case

\def\square{\hbox{\vrule\vbox{\hrule\phantom{o}\hrule}\vrule}}

\begin{document}

\author{Marek W. Gutowski\\
\medskip
{\small \sl Institute of Physics,
        Polish Academy of Sciences\\
        02--668 Warszawa, Al. Lotnik\'ow 32/46,
        Poland\\
        e-mail: gutow@ifpan.edu.pl}
}

\title{\Large\bf \uppercase{Amazing geometry of genetic space\\
        or\\
        are genetic algorithms convergent?}}
\maketitle

\begin{abstract}

    There is no proof yet of convergence of Genetic  Algorithms.  We  do 
 not supply it too. Instead, we present some thoughts and  arguments  to 
 convince the Reader, that Genetic Algorithms are essentially bound  for 
 success. For this purpose, we consider only  the  crossover  operators, 
 single- or multiple-point, together with selection procedure.

    We also give a proof that the soft selection is  superior  to  other 
 selection schemes.

\end{abstract}


\section{Introduction}
    We are using Genetic  Algorithms  (GA's)  for  solving  hard  global 
 optimization problems for at least three reasons:
\begin{itemize}
\item
    they are easy to implement in many computer languages,
\item
    they are applicable to problems, which cannot be easily, if at  all, 
 specified analytically as a set of closed-form formulas,
\item
    we believe, that their  inherent  intelligence  will  automatically, 
 i.e. with almost  no  programmer's  effort,  find  the  way  to  solve, 
 ''\/sufficiently well\/'', our difficult problems.
\end{itemize}
    The very idea of GA's, to simply mimic  the  Nature  \cite{Holland}, 
 belongs mostly to the  sphere  of  intuition,  and  is  almost  lacking 
 a~solid mathematical  background.  Indeed,  numerical  values  of  many 
 important ''\/tuning  parameters\/''  (mutation  rate,  probability  of 
 selection for reproduction, etc.) are largely selected on the  base  of 
 experience of other  people  solving  problems  similar  to  ours.  The 
 hypothesis   of   ''\/building   blocks\/''   appeared   false.   Other 
 investigations of GA's and their inner  working  are  rare.  We  simply 
 {\em believe\/}, that following the Nature's  paths  cannot  be  wrong. 
 But are we right? And, if so, why?

\section{Distance between parents and offsprings}

    It is easy to see, that after the crossover operation, the  distance 
 between resulting offsprings  is  identical  to  the  distance  between
 their   parents   \cite{matrix}.   Consider   a   pair   of    parents,
 $p_{a}=\left(p_{a}^{1},  p_{a}^{2},\  \ldots\,\  p_{a}^{N}\right)$  and
 $p_{b}=\left(p_{b}^{1},   p_{b}^{2},\    \ldots\,\    p_{b}^{N}\right)$
 consisting of  $N$  genes  each.  The  distance  between  them  may  be
 calculated in many ways, depending on metrics in use. In  the  simplest
 case, when each gene is just  a  binary  digit,  the  Hamming  distance
 ($d_{\rm H}\left(\cdot, \cdot\right)$)  is  perhaps  the  most  natural
 choice. This  simply  counts  the  number  of  bits  differing  on  the
 corresponding positions in the two given bit-strings.  It  is  obvious,
 that in this case
\begin{equation}\label{first}
    d_{\rm  H}\left(p_{a},  p_{b}\right)  =   d_{\rm   H}\left(   o_{a},
 o_{b}\right)
\end{equation}
    since the parents, $p_{a}$ and $p_{b}$, differ on exactly  the  same 
 positions as their offsprings, $o_{a}$ and $o_{b}$, do  ---  regardless
 of how many crossover points were used.

    When individual genes are more complex, i.e. when  they  consist  of 
 more bits (or, more often, are  the  symbols  drawn  from  finite  size
 alphabet(s)), or even when they are just the  real  numbers,  the  same
 remains true in any metrics  induced  by  $L_{p}$  norms.  Indeed,  the
 expression:
\begin{equation}
        d_{p}(p_{a}, p_{b}) = {\vert\!\vert p_{a}-p_{b}\vert\!\vert}_{p} 
        := \left[\sum_{k=1}^{N}\left|p_{a}^{k}-p_{b}^{k}\right|^{p}
        \right]^{\frac{1}{p}},\quad\ p=1,2,\ldots
\end{equation}
    has  to  be  equal  to  $d_{p}\left(o_{a},  o_{b}\right)$,  as   the 
 numerical components of the sum, shown above,  are  identical  in  both
 cases; even their order is preserved. The property (\ref{first})  holds
 also    for    less     frequently     used     norm     $L_{\infty}:$\
 ${\vert\!\vert{\mathbf x}\vert\!\vert}_{\infty} =
 \max_{k}\left|x_{k}\right|$.

\begin{figure}[ht]
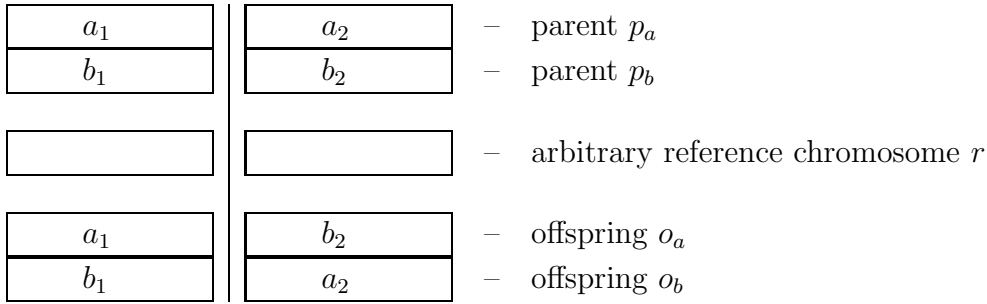

\begin{center}
\begin{tabular}{r|lcl}
\fbox{\phantom{BAB}$a_{1}$\phantom{$b_{1}$ABB}} &
\fbox{\phantom{BAB}$a_{2}$\phantom{$b_{2}$ABB}} & -- & parent $p_{a}$\\
\fbox{\phantom{BAB}$b_{1}$\phantom{$a_{1}$ABB}} &
\fbox{\phantom{BAB}$b_{2}$\phantom{$a_{2}$ABB}} & -- & parent
$p_{b}$\\
&&&\\
\fbox{\phantom{BAB$a_{1}$}\phantom{$b_{1}$ABB}} &
\fbox{\phantom{BAB$a_{2}$}\phantom{$b_{2}$ABB}} & -- & arbitrary reference
chromosome $r$\\
&&&\\
\fbox{\phantom{BAB}$a_{1}$\phantom{$b_{1}$ABB}} &
\fbox{\phantom{BAB}$b_{2}$\phantom{$a_{2}$ABB}} & -- & offspring
$o_{a}$\\
\fbox{\phantom{BAB}$b_{1}$\phantom{$a_{1}$ABB}} &
\fbox{\phantom{BAB}$a_{2}$\phantom{$b_{2}$ABB}} & -- & offspring
$o_{b}$
\end{tabular}
\end{center}
\caption{\sl Two parent chromosomes, $p_{a}$ and $p_{b}$,  arbitrary 
 reference chromosome, and the pair of offsprings, $o_{a}$ and  $o_{b}$.
 The integers, $a_{1}$, $a_{2}$, $b_{1}$ and $b_{2}$, denote numbers  of
 genes (bits), which are different in the respective parts  of  a  given
 chromosome and of the reference point in genetic space  ---  chromosome
 $r$.}
\label{cross}
\end{figure}

\medskip
    Consider now a~triangle in genetic space defined  by  vortices:  two 
 parent chromosomes, $p_{a}$  and  $p_{b}$  and  any  other  fixed,  but 
 otherwise arbitrary, reference point $r$. We will apply  the  crossover 
 operator to the  pair  $(p_{a},  p_{b})$,  obtaining  another  pair  of 
 chromosomes $(o_{a}, o_{b})$, as shown in Fig.~\ref{cross}.

    It is easy to verify (think of Hamming distance between  chromosomes 
 consisting of $1$-bit genes), that
\begin{equation}\label{circumference}
\left\{
\begin{array}{ccc}
        d_{\rm H}\left(p_{a}, r\right) &=& a_{1} + a_{2}\nonumber\\
        d_{\rm H}\left(p_{b}, r\right) &=& b_{1} + b_{2}\nonumber\\
        d_{\rm H}\left(o_{a}, r\right) &=& a_{1} + b_{2}\nonumber\\
        d_{\rm H}\left(o_{b}, r\right) &=& b_{1} + a_{2}
\end{array}
\right.
\end{equation}
    and,  after  adding  together  two  first  rows  of  the  above  and 
 comparing  the  result  with  the  sum  of  the  two   last   rows   of
 Eq.~\ref{circumference}, that
\begin{equation}
        d_{\rm H}\left(p_{a},r\right) + d_{\rm H}\left(p_{b},r\right) =
        d_{\rm H}\left(o_{a},r\right) + d_{\rm H}\left(o_{b},r\right) =
        a_{1} + a_{2} + b_{1} + b_{2}
\end{equation}
    In  remaining  cases,  with  discrete  or  continuous  genes,  other 
 measures of distance  between  them  may  be  used.  Looking  again  at 
 Fig.~\ref{cross}, we  can  conclude,  that  in  general  the  following 
 equality takes place:
\begin{equation}\label{general}
        d_{p}^{p}\left(p_{a}, r\right) + d_{p}^{p}\left(r, p_{b}\right) 
        = d_{p}^{p}\left(o_{a}, r\right) +
        d_{p}^{p}\left(r, o_{b}\right)
\end{equation}
    where  $p$  is  positive  and  finite\footnote{For  $p=\infty$   the 
 relation (\ref{general}) is usually false. Take $2$-genes  chromosomes:
 $p_{a}=\left(1, 5\right)$, $p_{b}=\left(2, 0\right)$,  and  $r=\left(0,
 0\right)$. Then $o_{a}=\left(1, 0\right)$,  $o_{b}=\left(2,  5\right)$,
 but  $d\left(p_{a},r\right)  +  d\left(p_{b},r\right)  =  7  \ne\  6  =
 d\left(o_{a},r\right) + d\left(o_{b},r\right)$.} integer,  and  $d_{p}$
 is a distance induced by $L_{p}$ norm.

    The relation (\ref{general}) may be extended even further, just  for 
 elegance, by adding to the left-hand  side  the  $p$-th  power  of  the 
 distance between parents and --- to the r.h.s --- $p$-th power  of  the 
 distance between offsprings, since, by virtue  of  (\ref{first}),  they 
 are equal to each other. Calling the sum of lengths of  the  triangle's 
 edges,  first  raised  to  the  fixed  integer  power  $p$,  the   {\em 
 generalized circumference\/}, we may express our result shortly as:
\begin{quote}
\sl
    The  generalized  circumference  of  the  triangle  made  of   three 
 chromosomes, remains unchanged when any two of  them  are  replaced  by
 their offsprings.
\end{quote}

\section{''\/Geometric\/'' conclusion and discussion}
    Recall that the chromosome  $r$  was  chosen  arbitrarily.  One  may 
 wonder what happens, if $r=x^{\star}$,\ i.e. when $r$ is the  searched, 
 still unknown, optimal chromosome ---  possibly  one  of  many  ---  in 
 a~genetic space.  If  this  is  the  case,  then  our  finding  may  be 
 expressed as follows:
\begin{quote}
\sl
    Since  the  sum  of  $p$-th  powers  of  distances  between   parent 
 chromosomes and the desired solution  is  conserved  by  the  crossover
 operator, then the offspring chromosomes cannot relocate too  far  from
 the optimal solution.
\end{quote}
\medskip\noindent{\em Proof:}

    Let the parents' distances from the solution  be  equal  to  $d_{1}$ 
 and $d_{2}$, both strictly positive, and the offsprings'  distances  --
 $d_{a}$ and $d_{b}$, respectively. We can write:
$$
        d_{1}^{p} + d_{2}^{p} = d_{a}^{p} + d_{b}^{p},\quad\ p=1, 2,
        \ldots\ <\infty
$$
or, equivalently
$$
        d_{b} = d_{2} \left[ 1 + \left(\frac{d_{1}}{d_{2}}\right)^{p} - 
        \left(\frac{d_{a}}{d_{2}}\right)^{p} \right]^{\frac{1}{p}},\quad\
        p=1, 2, \ldots\ <\infty
$$
    Depending on relation between $d_{a}$ and $d_{1}$ the value  of  the 
 expression appearing in the square  bracket  may  be  lower  than  $1$, 
 higher than $1$, or exactly equal to one. Respectively, we have
$$
\begin{array}{cccl}
        d_{a}<d_{1} &  \Rightarrow & d_{b}>d_{2}&\\
        d_{a}>d_{1} &  \Rightarrow & d_{b}<d_{2}&\\
        d_{a}=d_{1} &  \Rightarrow & d_{b}=d_{2}&\quad(\hbox{\rm
        ''\/rigid movement\/'')}
\end{array}
$$
    In words: if one of the offsprings  moves  further  apart  from  the 
 solution than one of its parents, then the second one  gets  closer  to
 the solution then their other parent. \square

\medskip{\em Comment}:
    In the degenerate case, when exactly one of the parents  is  already 
 an  optimal  solution,  it  may  happen  that  {\em  both\/}  offspring 
 chromosomes will be closer to the  solution  than  the  second  parent, 
 with none of them being the optimal  point,  which  quite  unexpectedly 
 appears to be a~repeller, rather than an attractor!

\medskip
    In conclusion: the outcome of the crossover operation may vary.  One 
 thing, however,  is  sure.  It  may  never  happen  than  {\em  both\/} 
 offspring chromosomes are located more far away from the solution  than 
 their more distant parent. On the other hand, there  is  no  guarantee, 
 that at least one of them gets closer to  the  desired  solution,  than 
 its ''\/better\/'' parent. Nevertheless, {\bf at  least  one  of  newly 
 created individuals is equally or less distant from the  solution  than 
 its ''\/worse\/'' parent\/}. We use quotation marks, since  in  reality 
 the chromosomes located closer to the optimal one (the  ''\/better\/'') 
 need not to be better fitted. This is most easily seen in  cases,  when 
 genotypes arbitrarily close to the solution are  unacceptable  at  all, 
 for example due to violation of constraints. So  the  really  important 
 question is: {\bf Which one of the two  offsprings  is  closer  to  the 
 solution?}\ Judgment based on their fitness alone  cannot  be  regarded 
 as reliable or conclusive.

\begin{table}[h]
\begin{center}
    \caption{\sl Possible  outcomes  of  the  crossover  operation.  The 
 positions of symbols ($p$ -- parent, $o$ -- offspring) are  meaningful:
 the more to the right -- the bigger is the distance  from  the  desired
 solution.  Omitted  are  the  cases,  when  distances   either   remain
 unchanged or coincide.}
\label{tabelka}
\begin{tabular}{rcr}
\hline
{\sl No.} & {\sl configuration} & {\sl comments}\\
\hline\hline
1 & $oopp$ & impossible\\
2 & $opop$ &\\
3 & $oppo$ &\\
4 & $poop$ &\\
5 & $popo$ &\\
6 & $ppoo$ & impossible\\
\hline
\end{tabular}
\end{center}
\end{table}

    We shortly summarize  all  interesting  outcomes  of  the  crossover 
 operation in Tab.~\ref{tabelka}. Analyzing its contents  we  see,  that 
 the symbol ''\/$o$\/'' can be found at  advantageous  position  exactly 
 $6$ times, while only twice on disadvantageous one. Does it mean,  that 
 the odds for selecting ''\/proper\/'' offsprings, i.e.  to  improve  at 
 least one trial solution, are $6:2$? The answer would be  positive,  if 
 the  events  $2$---$5$  occurred  with  equal  probability,   what   is 
 unlikely. On the other hand, if only the case $5$  (the  worst)  occurs 
 again and again, then the random, unbiased  selection  of  one  of  the 
 offsprings, would give us {\em exactly\/} (!)  $50$\%  chance  to  move 
 closer to the reference chromosome $r$. This means, that  in  practice, 
 the chances for improvement can be even higher than  $\frac{1}{2}$;  we 
 will prove that, rigorously, in the following section.

\medskip\noindent{\bf Important note:}
    We have to carefully distinguish  between  continuous  and  discrete 
 case. In discrete genetic  space  the  only  convergent  sequences  are 
 constant sequences. This is because there are no elements  of  discrete 
 genetic  space,  which  would  be  located  arbitrarily  close  to  any 
 existing chromosome, the  optimal  one  in  particular.  Therefore  the 
 notion of convergence is sensible and usable only in continuous cases.

\medskip
    On the other hand, since $r$ is arbitrary, then it may have  nothing 
 to do with the optimal solution. That is why  the  entire  evolutionary 
 process may not converge at  all  without  additional  driving  forces, 
 other than the actions of crossover operators.

    As we will show now, the key to the question of convergence  is  the 
 selection process -- the practical realization of  the  Darwinian  rule 
 of  evolution,  {\em  survival  of  the  fittests\/},   understood   in 
 a~probabilistic sense rather than an absolute rule.

\medskip

\section{Chances for success}
    The following text is based on the  problem  stated  and  solved  by 
 Lata{\l}a in {\em Delta\/} \cite{delta} -- a~popular Polish monthly  on 
 mathematics, physics and  astronomy,  targeted  mainly  at  high-school 
 students. The problem and its solution  are  freely  rephrased  by  the 
 current author.

\medskip\noindent
    Problem:
\begin{quote}
\sl
    Find the winning strategy in the following game:\\
    Looking at an integer number, randomly chosen from two such  numbers 
 written down by our opponent, guess whether the other (unknown)  number 
 is higher or lower. The two numbers in question are distinct.  We  win, 
 when our guess is correct, otherwise we loose.
\end{quote}

\pagebreak[4]
\noindent
Solution:
\begin{quote}
\sl
    Use arbitrary, strictly increasing, sequence of numbers 
    $\left\{c_{k}\right\}_{k=-\infty}^{\infty}$, each belonging to the
    (open) interval $\left(0, 1\right)$, for example
    $c_{k}=\frac{1}{2} + \frac{{\rm arc}\tan k}{\pi}$. When the selected
    (known) number is equal to $k$, then with probability $c_{k}$
    ''\/guess\/'', that the other (unknown) number is {\em lower\/}, or,
    with probability $1-c_{k}$, that it is {\em higher\/}.
\end{quote}

\medskip\noindent
    It is obvious, that this strategy should work equally well not  only 
 for unknown integer numbers, but also when the numbers are  drawn  from 
 any countable subset of reals. But why does it work at all?

\medskip
    Let $p_{m,n}$ denotes the probability, that the  numbers  chosen  by 
 our opponent are $m$ and $n$, and that $m>n$. The probability, that our
guess is correct, may be written as
\begin{equation}\label{solve}
        \sum_{m,n\in\hbox{\mm Z},\ m>n} p_{m,n} \left[ \frac{1}{2}c_{m} 
        + \frac{1}{2} \left( 1 - c_{n}\right) \right] = \frac{1}{2} +
        \frac{1}{2} \sum_{m,n\in\hbox{\mm Z},\ m>n} p_{m,n} \left(c_{m}
        - c_{n} \right)
\end{equation}
    where   \hbox{\mm   Z}   is   the    set    of    integers.    First 
 ''\/$\frac{1}{2}$\/'' in the r.h.s. of  (\ref{solve})  comes  from  the 
 fact,  that  $\sum  p_{m,n}=1$.  The  second  component   is   strictly 
 positive, since $c_{m}>c_{n}$ for  arbitrary  $m>n$  (as  the  sequence 
 $\left\{c_{k}\right\}$ is strictly increasing)  and  at  least  one  of 
 $p_{m,n}$ is greater than zero.  In  conclusion:  our  chances  to  win 
 always {\em exceed\/} $50$\%. This wouldn't  be  so,  if  the  sequence 
 $\left\{c_{k}\right\}$  was  not  strictly  increasing   --   in   such 
 circumstances our chances to win could be estimated only  as  {\em  not 
 less\/} than $\frac{1}{2}$. Let us note, that nothing  certain  can  be 
 said about {\bf how much\/} our chances  to  win  exceed  $50$\%.  They 
 will peak, if all  the  differences  $c_{m}-c_{n}$  are  maximized,  at 
 least  those  of  them,  which  ''\/meet\/''  non-zero  $p_{m,n}$'s  in 
 formula (\ref{solve}).  Unfortunately,  we  know  nothing  ahead  about 
 probabilities $p_{m,n}$'s.

\medskip
    How is the  above  problem  related  to  Genetic  Algorithms?  Quite 
 simply: the  sequence  $\left\{c_{k}\right\}$  should  be  regarded  as 
 a~tool to convert the value of fitness to probability of selection. The 
 superiority of the  soft  selection,  realized  with  such  a  sequence 
 $\left\{c_{k}\right\}$, over the hard selection  schemes,  is  evident. 
 In the case of soft selection, our chances to win (i.e. to improve  the 
 objective by selecting a~better offspring for further  processing)  are 
 always higher than chances for  failure.  On  the  contrary,  the  hard 
 selection\footnote{i.e.   $c_{n}\equiv\   0$   for    $n<n_{0}$,    and 
 $c_{n}\equiv 1$ otherwise.  If  so,  then  $c_{m}-c_{n}$  in  r.h.s  of 
 (\ref{solve}) is necessarily equal to  zero  for  many  pairs  $(m,n),\ 
 m>n$. For those pairs $(m,n)$,  for  which  $c_{m}-c_{n}=1$,  in  turn, 
 $p_{m,n}$ may be equal to zero --  corresponding  to  the  pairs  never 
 produced by our malicious (smart?)  opponent.}  implies  that,  in  the 
 unlucky event, both chances can be equal to each other.

    The hard selection scheme can be  considerably  improved  to  become 
 comparable with the soft selection. It is enough to select  the  number 
 $n_{0}$ (see footnote) as any average of fitnesses of  all  individuals 
 present in the previous generation. This  trick  should  work  best  in 
 cases, when our opponent -- the objective  function  --  produces  only 
 a~few discrete values.

    The  difference  seems  rather  subtle:  sharp  versus   not   sharp 
 inequality. But let us recall the brutal practice  of  citizens  of  an 
 ancient Greek  city  of  Sparta.  In  strive  to  have  only  excellent 
 warriors as their descendants, they used to  physically  eliminate  all 
 ''\/defective\/'' newborns. Did they succeed?

\section{More on selection}
    Consider the objective function with  many  local  extrema  of  very 
 similar fitness value, yet  having  exactly  one  global  optimum.  The 
 evolving population will  sooner  or  later  split  into  many  loosely 
 connected clusters, concentrated around those extrema. To discover  the 
 true, global optimum,  we  need  the  ability  to  correctly  rank  the 
 individuals with very close values of  their  fitness.  Only  then  the 
 ''\/useless\/'' individuals, located around  local  extrema,  would  be 
 extinct. Therefore, in particular computer  implementation,  not  every 
 kind of average used as threshold  for  hard,  stepwise  selection,  is 
 equally good. To increase our chances for success, and  accelerate  the 
 convergence    as    well,    we    should    apply    the     sequence 
 $\left\{c_{k}\right\}$, or its continuous counterpart --- which may  be 
 selected individually for each new generation --- in such  a~way,  that 
 it changes most significantly around majority of fitness values  across 
 the population. When searching for maximum, the  following  simple  and 
 numerically plausible transformations from $fitness$  to  $probability\ 
 of\ selection$,\ often called {\em scaling of the fitness  function\/}, 
 satisfy this requirement:
\begin{equation}\label{scale}
        c_{k} = \frac{1}{2} + \frac{1}{\pi} {\mathrm arc}\tan 
        \frac{2\left(k-f_{1/2}\right)}{f_{3/4}-f_{1/4}}\quad\ {\mathrm
        or}\quad\ c_{k}^{\prime} = \frac{1}{2} + \frac{1}{2}
        \tanh\frac{2\left(k-f_{1/2}\right)}{f_{3/4}-f_{1/4}}
\end{equation}
    with the first  choice  being  definitely  softer.  The  subscripted 
 constants $f_{\alpha}$ denote  respective  quantiles  (more  precisely: 
 quartiles) of the fitness distribution across the  current  population, 
 with  $f_{1/2}$  being  the  median.  Put  unity\footnote{The   numbers 
 appearing in both denominators need not to be computed very  precisely. 
 Our choice is dictated by purely numerical reasons: neither the poorest 
 individuals are neglected, nor the best fitted ones have the  guarantee 
 to  be  selected.}  into  the  denominator  when  $f_{3/4}-f_{1/4}=0$.\ 
 Replace summation in (\ref{scale}) with subtraction, when searching for 
 minimum.

\medskip
    It is clear, that GA's can be most  effective  for  objectives,  for 
 which only a very limited information is available, namely nothing  but 
 fitness values computed for every member  of  the  population,  usually 
 only the last  one.  Their  ability  to  quickly  detect  and  then  to 
 concentrate in the interesting parts of the  search  space  makes  them 
 clearly superior to generic Monte Carlo approach, which waste time  for 
 uniform and fruitless  exploration  of  other  regions.  The  above  is 
 certainly true for objectives, which are at least piecewise  continuous 
 and have no singularities. For such a broad  class  of  problems,  with 
 chromosomes coded in a~natural way\footnote{By natural coding  we  mean 
 such a~mapping of continuous unknowns to genes representing them, which 
 is strictly monotonous, and therefore invertible.}, the  {\bf  stopping 
 rules\/}   should   be   based   on   compactness   of   the   evolving 
 po\-pu\-la\-tion, paying only  little  attention  to  the  behavior  of 
 fitness. The evolutionary process should be continued as  long  as  the 
 volume of the search space  occupied  by  ''\/better  half\/''  of  the 
 population still decreases. One should be  aware,  however,  that  this 
 strategy will fail for objectives with more than one global optimum  or 
 when the unique extremum is degenerate (flat, improper), i.e.  consists 
 of more than a single point, either in reality or due to roundings.  If 
 this is the case, then careful analysis of the last generation  may  be 
 helpful.

    For discrete problems (with  integer  and  maybe  boolean  variables 
 present) the notion of continuity does not apply, so  the  task  is  to 
 efficiently find the acceptable  solution  {\em  without\/}  performing 
 exhaustive search.  It  can  be  shown  \cite{nova},  that  for  purely 
 discrete  problems  we   need   ${\mathcal   O}\left(N^{\frac{3}{2}}\ln 
 N\right)$ evaluations of the objective instead of $2^N$,  as  necessary 
 and required by the exhaustive search. $N$ is the number of  bits,  not 
 unknowns,   in   a~single   chromosome.   To    be    precise:    after 
 $N^{\frac{3}{2}}\ln N$ evaluations of the objective,  the  chance  that 
 the best so far chromosome is  separated  no  more  than  $1$  unit  of 
 distance (in Hamming sense) from  the  optimal  one,  are  higher  than 
 $50$\%. No well justified stopping rules  can  be  given  for  discrete 
 case.

    Mixed problems, involving real {\em  and\/}  integer  unknowns,  are 
 even harder to analyze. From the formal point of  view,  such  problems 
 may be regarded as large, but finite, sets of continuous problems.

\section{Summary}
    We have shown, that Genetic Algorithms are bound  for  success.  The 
 chances for improvement are always higher than for lack of it,  if  the 
 selection of parents  is  performed  either  in  a~soft,  or  hard  but 
 adaptive,  manner.  This  is  a   very   general   result,   completely 
 independent  on  the  optimization  problem  under  study.  It  applies 
 equally well to discrete, continuous and mixed optimization problems.

    As we can see, the quite high  chances  of  Genetic  Algorithms  for 
 success are strictly related to their property  of  not  rejecting  nor 
 ignoring the ''\/bad\/'' trial points in the  search  space.  Contrary, 
 the  rigorous,  deterministic  search  methods  are  simply  unable  to 
 ''\/jump over\/''  the  barrier  surrounding  even  the  single  global 
 minimum, if started too far from the solution.

    Our result is of stochastic nature rather than  deterministic.  This 
 may mean in practice, that we may be  unable  at  all  to  improve  the 
 already known, approximate solution of our particular problem.  Nothing 
 can be said how quickly we will arrive at any improvement. This may  be 
 significantly  influenced  by  other  components  of   GA's:   mutation 
 operators, population size  and  numerical  values  of  various  tuning 
 parameters, not on the selection scheme or crossover mechanism alone.

\section*{Acknowledgment}
    This work was done as a part of author's statutory activity  at  the 
 Institute of Physics, Polish Academy of Sciences.

\end{document}